%% file: main.tex
\documentclass[a4paper]{article}

\usepackage{INTERSPEECH2020}

\usepackage{amsmath}
\usepackage{amsfonts}
\usepackage{graphicx}
\usepackage{makecell}
\usepackage{soul,color}
\usepackage[ruled,vlined]{algorithm2e}
\usepackage[inline]{enumitem}
\usepackage{subfigure}
\usepackage{multirow}
\usepackage{multicol}
\usepackage{pifont}
\usepackage{balance}
\newcommand{\parspace}{\vspace{+0.35em}}

\newcommand{\cmark}{\ding{51}}%
\newcommand{\xmark}{\ding{55}}%

\DeclareMathAlphabet{\mymathbb}{U}{BOONDOX-ds}{m}{n}

\title{Iterative Compression of End-to-End ASR Model using AutoML}

\name{Abhinav Mehrotra$^{1}$, \L ukasz Dudziak$^{1}$, Jinsu Yeo$^{2}$, Young-yoon Lee$^{2}$, Ravichander Vipperla$^{1}$, Mohamed S. Abdelfattah$^{1}$, Sourav Bhattacharya$^{1}$, Samin Ishtiaq$^{1}$, Alberto Gil C. P. Ramos$^{1}$, SangJeong Lee$^{2}$, Daehyun Kim$^{2}$, Nicholas D. Lane$^{1,3}$} 
\address{
    $^{1}$ Samsung AI Center, Cambridge, UK \\
	$^{2}$ On-device Lab, Samsung Research, Seoul, South Korea \\
	$^{3}$ University of Cambridge, UK 
	}
	
\email{}

\definecolor{mygreen}{rgb}{0,0.6,0}
\definecolor{mygray}{rgb}{0.5,0.5,0.5}
\definecolor{mymauve}{rgb}{0.58,0,0.82}

\begin{document}

\maketitle

\begin{abstract}
Increasing demand for on-device Automatic Speech Recognition (ASR) systems has resulted in renewed interests in developing automatic model compression techniques. Past research have shown that AutoML-based Low Rank Factorization (LRF) technique, when applied to an end-to-end Encoder-Attention-Decoder style ASR model, can achieve a speedup of up to 3.7$\times$, outperforming laborious manual rank-selection approaches. However, we show that current AutoML-based search techniques only work up to a certain compression level, beyond which they fail to produce compressed models with acceptable word error rates (WER). In this work, we propose an iterative AutoML-based LRF approach that achieves over 5$\times$ compression without degrading the WER, thereby advancing the state-of-the-art in ASR compression.

\end{abstract}
\noindent\textbf{Index Terms}: ASR Compression, AutoML, Reinforcement Learning

\input{intro}

\input{iterative_search}

\input{results}

\section{Conclusions}
\label{sec:conclusion}
\input{conclusions}


\bibliographystyle{IEEEtran}
\balance
\bibliography{main}

\end{document}

%% file: intro.tex
\section{Introduction}
\label{sec:intro}

Rapid technological improvements in deep learning-based acoustic modeling, language modeling, and noise-resilience techniques resulted in a drastic drop in the WER of modern ASR systems. Evidently, ASR is serving as the backbone in audio-based input modality on a variety of devices including mobile phones, smart speakers, and IoT appliances. Due to existing data security and privacy concerns in cloud-based ASR systems, a clear shift in preference towards on-device deployment of the state-of-the-art ASR models is emerging. Mobile and IoT devices, however, suffer from a limited resource budget and require efficient deployment of ASR models with significantly lower memory, compute and power demands.

\parspace

\noindent
Popular techniques in reducing resource demands of well-trained and parameter-heavy models include LRF, pruning, and reduced-precision representations~\cite{pang_compression_2018, hinton_kd_2015}. In our previous work, we have shown that automated Reinforcement Learning (RL)-based search can be applied to identify low ranks in LSTM weight-matrices, which allowed for a $1.23\times$ relative speedup gain over a manual search procedure~\cite{shrinkml2019}. However, we observe that the conventional AutoML search fails to find ranks that manifest higher compression ratio (e.g., $\geq 3.7\times$), without degrading the WER. 
This is due to inability of RL to differentiate between better and worse choices since most of the visited points result in similar rewards.
Alternatively, iterative approaches have been successfully used to manually compress image recognition models, often delivering better results than their one-shot counterparts~\cite{liu_learning_2017,molchanov_pruning_2017,frankle_lottery_2018,gao_dynamic_2019}. 
\parspace


\begin{figure}[t!]
  \centering
  \includegraphics[width=0.8\linewidth]{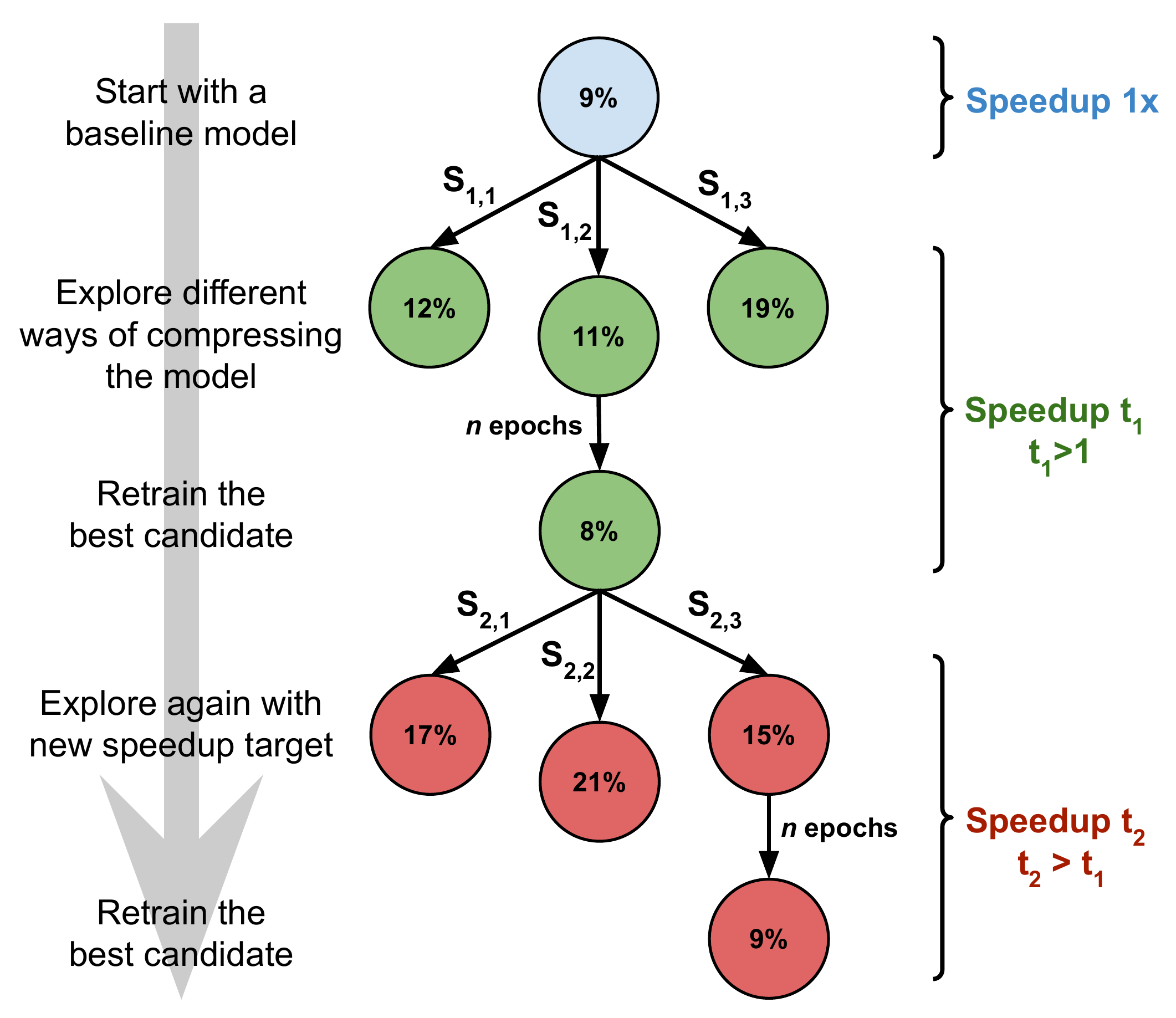}
  \caption{A high-level overview of the proposed method. Percentages in nodes are example word error rates for different models. $s_{n,i}$ values represent different schemes of compressing a model to achieve target speedup $t_n$.}
  \label{fig:iter_compression}
  \vspace{-5mm}
\end{figure}

\noindent
In this work, we propose a unified approach by combining iterative compression with AutoML-based rank searching to push the boundaries of ASR model compression. We present an RL-based iterative search that performs incremental compression by following a sequence of increasing speedup targets (\textit{trajectory}) to reach the desired final speedup, while maintaining the WER.
The basic idea behind our method is illustrated in Figure~\ref{fig:iter_compression}.
The main goals are: (i) obtain smaller models than the one achievable by current AutoML-based one-shot compression, while maintaining low WER; and (ii) understand if there is a fundamental difference between iterative and one-shot LRF-based compression techniques. 
\parspace

\noindent
We present an extensive set of experiments by considering a state-of-the-art E2E ASR model and show that the proposed iterative search can achieve at least $5\times$ compression ratio, while producing a $7\%$ relative gain in WER. 
Compared to the current state-of-the-art one-shot AutoML compression the results show $1.35\times$ relative gain in speedup without degrading the model's accuracy. 
Furthermore, we investigate the impact of different trajectories on achieving a $5\times$ compression ratio with our iterative compression approach. Our results suggest that taking incremental compression steps, while performing iterative search, helps in discovering better compression ranks. 
Finally, we show that once the optimal ranks are found, they can be applied to the baseline model in an one-shot manner to obtain a compressed model without any loss in WER.
However, the question, {\em how to find those ranks without intermediate retraining}, remains an open problem.

%% file: iterative_search.tex
\begin{algorithm}[t!]
\begin{footnotesize}
\LinesNumbered
\SetAlgoLined
\KwIn{(i) Trajectory $[t_i]_{i=1}^K,\, \mbox{where } t_i \in \mathbb{R}^{+}$, (ii) parameter matrices $\mathbb{M} = \{M_i\}_{i=1}^L$.}
\KwOut{Compressed and retrained model}
\For{$i\leftarrow1$ \KwTo K}{
    $S \leftarrow \text{construct\_search\_space}(\mathbb{M}, t_i)$ 
    $s \leftarrow \text{automatic\_rank\_search}(S, \mathbb{M}, t_i)$ 
    \For{$j\leftarrow1$ \KwTo L}{
        $U, \Sigma, V^T \leftarrow LRF(M_j)$ 
        
        $\hat{M}_j \leftarrow U_{[:s_j]} \Sigma_{s_j} V_{[s_j:]}^T$ 
        
        $M_j \leftarrow \hat{M_j}$
    }
    \If{i == K}{
        $s^{*} \leftarrow s$
    }
    Retrain the model using $\mathbb{M}$ for initialization
}

\For{$j\leftarrow1$ \KwTo L}{
    $U, \Sigma, V^T \leftarrow LRF(M_j)$ 
    
    $M_j \leftarrow  (U_{[:s^*_j]}, \Sigma_{s^*_j}V^T_{[s^*_j:]})$
}
Retrain the model using $\mathbb{M}$ for initialization

\caption{Iterative compression with AutoML.}
\label{alg:iter}
\end{footnotesize}
\end{algorithm}

\section{AutoML-based Iterative Compression}
\label{sec:iter}

\noindent
\noindent
\textbf{Automatic Compression Scheme.} Given a layer parameter matrix $M$, we use LRF as our main compression method~\cite{xue_restructuring_nodate,sainath_low-rank_2013,sourav} and approximate 
$M$ using its truncated singular value decomposition (SVD), i.e., $M = U \Sigma V^T \approx U_{[:k]}\Sigma_{k}V^T_{[k:]} = \hat{M}$.
Given a ASR model, we compress all parameter matrices independently by choosing different ranks and call the set of selected ranks across all layers a \textit{compression scheme} ($s$). We further define \textit{speedup} (or compression ratio) of the compressed model as: 
\begingroup
\setlength\abovedisplayskip{0pt}
\setlength\belowdisplayskip{4pt}
\[\text{speedup} = \frac{\text{FLOPS of baseline}} {\text{FLOPS of compressed model}}\]
\endgroup
We automatize the process of searching for a good compression scheme by using an algorithm based on REINFORCE~\cite{zoph_neural_2016} with two reward functions, $\mathcal{R}$ and $\mathcal{R}_{v}$, used to optimize for lower WER and smaller model size simultaneously. Further details can be found in~\cite{shrinkml2019}.


\noindent
\textbf{Iterative Compression.} Iterative model compression is performed by repeatedly applying AutoML-based rank-search for a {\em trajectory}, followed by retraining the model with the identified best compression scheme. A trajectory ($t = [t_1, t_2, ... t_K]$) is defined as a sequence of increasing intermediate speedup goals to reach the final target speedup. For instance, example trajectories for an overall $3\times$ speedup goal can be $[2\times, 3\times]$ and $[1.5\times, 2\times, 2.5\times, 3\times]$, resulting in $2$ and $4$ iterations of AutoML search respectively. Note that \textit{one-shot AutoML-based search} is a special case of the iterative searching, where the trajectory consists of a single step, e.g., $[3\times]$. 
\parspace

\noindent
During each step within a trajectory, the best compression scheme $s$ found by the automated search is used to approximate individual parameter matrices of the model, i.e., $\hat{M}_i = U_{[:k]}\Sigma_{k}V^T_{[k:]}$, where $i$ is the index over all layers. These approximations are then used as the initialization weights for layers during model retraining for a fixed number of epochs. To leverage computational and memory gains, we only store factorized components $U_i, V^T_i$ for the final model identified at the end of a trajectory search.
However, we keep using the full size matrices $\hat{M_i}$ during intermediate iterations to give the training algorithm more capacity in recovering from degraded WER due to low-rank approximations.
The overall iterative search procedure is described in Algorithm~\ref{alg:iter}.

\parspace

\noindent
At the first search step of a trajectory, we identify good range of ranks (for a specific layer) based on the cumulative eigen-value energy range, for obtaining the required speedup. 
For the subsequent steps we use the same energy range when the $\Delta t_i$ (i.e., the speedup increase from the previous step) is similar to $\Delta t_{i-1}$. However, we lower (or increase) the range to be more aggressive (or moderate) for high (low) values of $\Delta t_{i} / \Delta t_{i-1}$. 
For example, energy range considered during the search for $2\times$ speedup, is reused when we perform the search for $3\times$ speedup in the next step. However, we lower the energy range if we perform the search for $4\times$ speedup after $2\times$ compression. 





%% file: results.tex
\section{Evaluation}
\label{sec:exp}

\begin{figure}[t!]
\centering
  \includegraphics[width=0.75\linewidth]{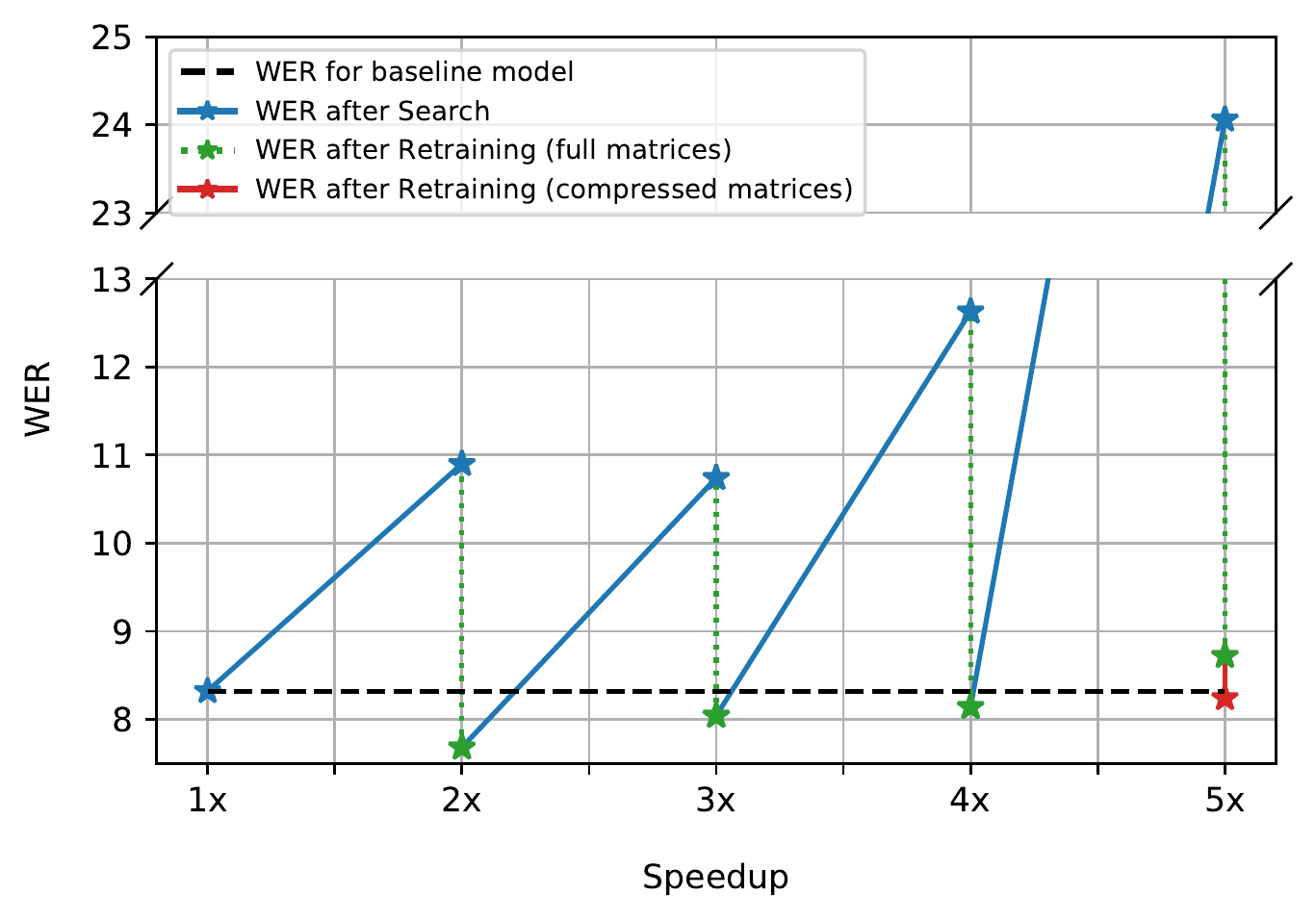}
\vspace{-3mm}
  \caption{WER for MoChA model at each step of iterative compression approach. The solid lines indicate the change in WER during each compression step, and the dotted lines show the gain in performance after retraining. The red line indicate the performance gain after retraining target compressed model.}
  \label{fig:wer}
\vspace{-5mm}
\end{figure}

\begin{table*}[t]
\centering
\begin{footnotesize}
\begin{tabular}{ l l c c c c c c c } 
\hline
\textbf{Part} & \textbf{Layer} & \textbf{Search} & \textbf{Orig. dimensions} & \textbf{Orig. FLOPS [M]} & \textbf{Rank} & \textbf{New FLOPS [M]} & \textbf{Speedup} \\ 
\hline
\hline
\multirow{6}{*}{Encoder} & LSTM 0 & \cmark & 40$\times$4096, 1024$\times$4096 & 0.16, 4.29 & Full, 105 & 0.16, 0.55 & 1x, 7.8x \\  
 & LSTM 1 & \cmark & 1024$\times$4096, 1024$\times$4096 & 4.29, 4.29 & 72, 85  & 0.37, 0.44 & 11.4x, 9.6x \\  
 & LSTM 2 & \cmark & 1024$\times$4096, 1024$\times$4096 & 4.29, 4.29 & 82, 63  & 0.42, 0.32 & 10.0x, 13.0x \\  
 & LSTM 3 & \cmark & 1024$\times$4096, 1024$\times$4096 & 4.29, 4.29 & 102, 81 & 0.52, 0.41 & 8.0x, 10.1x \\  
 & LSTM 4 & \cmark & 1024$\times$4096, 1024$\times$4096 & 4.29, 4.29 & 93, 103 & 0.48, 0.53 & 8.8x, 8.0x \\  
 & LSTM 5 & \cmark & 1024$\times$4096, 1024$\times$4096 & 4.29, 4.29 & 80, 82  & 0.41, 0.42 & 10.2x, 10.0x \\  
\hline
\multirow{4}{*}{Attention} &enc. ctx & \cmark & 1024$\times$1024 & 1.05 & 16 & 0.03 & 32.0x \\  
 & chunk enc. ctx    & \cmark & 1024$\times$1024 & 1.05 & 24 & 0.05 & 21.3x \\  
 & dec. trans.       & \cmark & 1000$\times$1024 & 1.02 & 15 & 0.03 & 33.7x \\  
 & chunk dec. trans. & \cmark & 1000$\times$1024 & 1.02 & 17 & 0.03 & 29.8x \\  
\hline
Decoder & kernel & \cmark & 2645$\times$4000 & 10.58 & 130 & 0.86 & 12.2x \\  
\hline
\multirow{3}{*}{Output} & readout & \xmark & 2645$\times$1000 & 2.64 & N/A & 2.64 & 1x \\
 & output prob. & \xmark & 500$\times$10025 & 5.01 & N/A & 5.01 & 1x \\
 & embedding    & \xmark & 10025$\times$621 & 0    & N/A & 0     & 1x \\
\hline
\hline
\multicolumn{4}{r}{\multirow{2}{*}{\textbf{Overall:}}} & \multirow{2}{*}{\textbf{63.96}} & \multirow{2}{*}{N/A} & \multirow{2}{*}{\textbf{12.73}} & \multirow{2}{*}{\textbf{5.0x}} \\
\multicolumn{4}{r}{} & & & & \\
\hline
\end{tabular}
\end{footnotesize}
\vspace{1mm}
\caption{Break-down of the Base--2--3--4--5x compression scheme. LSTM layers contain two matrices (for projecting input and recurrent state) hence two values are reported. Layers which do not contribute much to the running-time were omitted. However, we included the embedding layer due to its considerable size (even though it has marginal impact on run-time). 
}
\label{tab:breakdown}
\vspace{-10mm}
\end{table*}

\paragraph*{Model Architecture.} We use an end-to-end encoder-attention-decoder ASR model~\cite{asru1,asru2,Zeyer2018ImprovedTO} with unidirectional LSTMs (unit size 1024) and monotonic chunk-wise attention (MoChA)~\cite{chiu2018monotonic} as our baseline. We identified 11 layers consisting of total 17 matrices which could be compressed and considered 5 different compression levels for each of them, which constitutes a search space of $5^{17} \approx 762\times 10^9$ possible combinations.

\vspace{-0.4cm}
\paragraph*{Model Training.} The model is trained on LibriSpeech dataset~\cite{Panayotov2015Libri} using MFCC features as input and operates on a vocabulary of 10,000 byte-pair-encoded sub-word units~\cite{Sennrich2016bpe, isck1}.
The framework used for training and evaluation of word error rate (WER) is RETURNN~\cite{zeyer2018returnnAcl} with tensorflow~\cite{tensorflow2015-whitepaper} backend.

\vspace{-0.4cm}
\paragraph*{Baseline Compression Approach.} We use the state-of-the-art AutoML-based one-shot compression~\cite{shrinkml2019} as the baseline for evaluating the performance of our approach.

\subsection{Defining Trajectories}
\label{sec:trajectories}

To perform iterative compression, we need to define a trajectory that informs the iterative search algorithm about the intermediate speedup through which it should traverse. Since a trajectory is based on the target speedup, we performed initial experiments to explore the maximum possible compression. Based on our empirical results, we found that some layers of the model used in the experiment reach the maximum compression with the target speedup of 5x (discussed in Section~\ref{sec:iter_compression}). Therefore, to evaluate our iterative compression approach we defined a series of trajectories to obtain 5x speedup, which is $1.35\times$ improvement over the state-of-the-art AutoML-based one-shot approach~\cite{shrinkml2019}. Conclusively, we define the following trajectories:
\begin{enumerate*} [label=(\roman*)]
\item Base-2x-5x
\item Base-2x-3x-5x
\item Base-2x-3x-4x-5x
\item Base-3x-5x
\item Base-3x-4x-5x
\item Base-4x-5x. 
\end{enumerate*}
Note that we used the increment of speedup during each step of a trajectory as a factor of 1x, but this could be any real number as per the requirements.

\begin{figure}[t!]
\centering
  \includegraphics[width=0.825\linewidth]{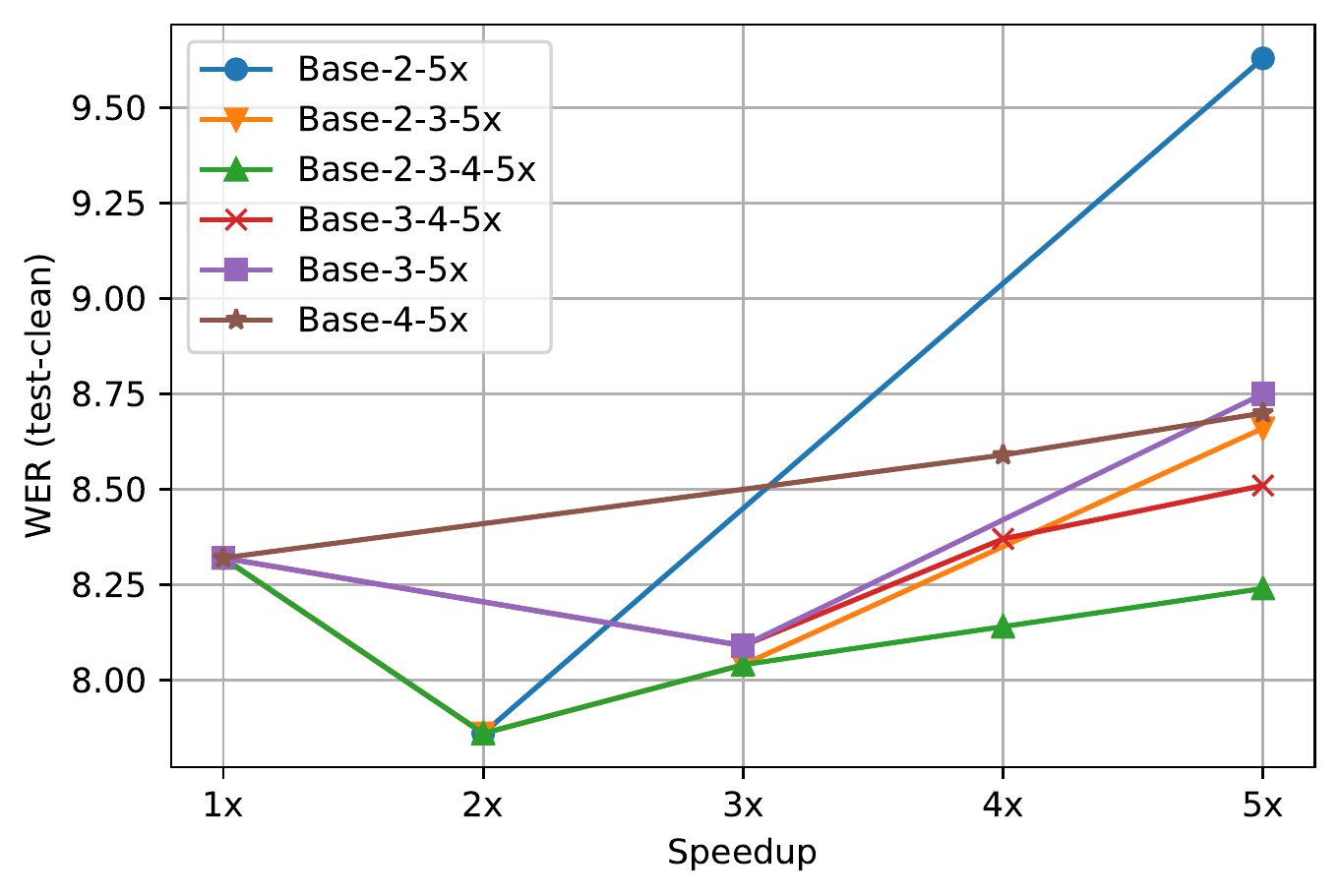}
\vspace{-4mm}
  \caption{WER for MoChA model for $5\times$ compression with different granularities of steps.}
  \label{fig:trajectories}
\vspace{-6mm}
\end{figure}

\subsection{Iterative Model Compression}
\label{sec:iter_compression}


To perform an iterative search for a trajectory, energy range are explored between 0.3 to 0.99, such that they could find good ranks to compress the model satisfying the speedup threshold of the first step of the trajectory. 
These ranks are used to compress the model, which is then retrained\footnote{Through the initial experiments we found that around 50 epochs were required to recover the model performance after applying SVD while increasing speedup by 1x.} to recover any drop in WER. This model is then used as a base to perform the compression search for the next step of the trajectory. 
The processes is continued with the compression-retraining phases repeated until all steps of the trajectory are processed. 
\parspace

\noindent
Following this methodology, we performed the iterative search for all trajectories defined in Section~\ref{sec:trajectories}. The results show that \textit{Base-2$\times$-3$\times$-4$\times$-5$\times$} trajectory was able to find the best ranks for 5.0x compression along with a 7\% relative gain in accuracy (i.e., reduced WER). Thus, our approach outperforms the current AutoML-based one-shot compression with $1.35\times$ gain in speedup. 
\parspace

\noindent
The entire process of the best trajectory (i.e., Base-2$\times$-3$\times$-4$\times$-5$\times$) is depicted in Figure~\ref{fig:wer}, with solid and dotted lines representing search/compression and retraining steps respectively, and Table~\ref{tab:breakdown} shows a breakdown of the obtained compression scheme. We observe that at 5x compression our model becomes mainly limited by the size of the layers which were not subject to the search, with the other parts being compressed well beyond 5x (i.e., around 13x) to balance them out. Interestingly, layers related to attention turned out to be the most compressible with around 28x speedup.

\begin{figure}[t!]
\centering
  \includegraphics[width=0.95\linewidth]{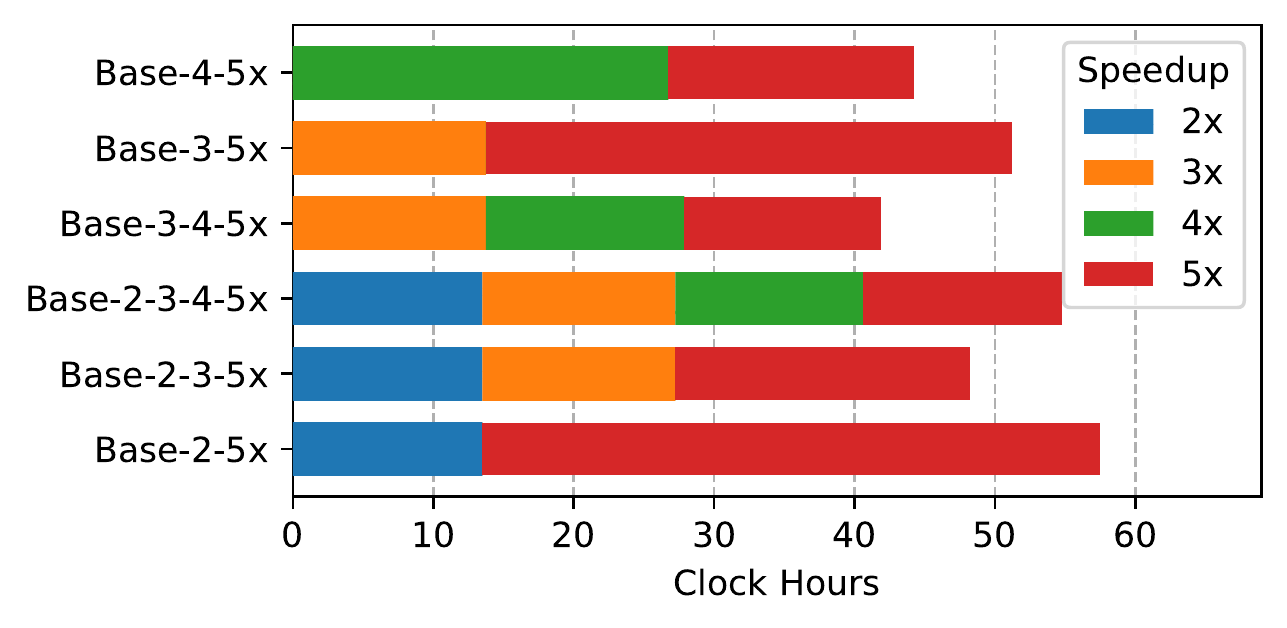}
\vspace{-4mm}
  \caption{Computation time for 5x compression of MoChA model with different granularity of step. Each color indicates the specific speedup of a step and the length represents the clock hours consumed during search for that step (excluding training time).}
  \label{fig:trajectories_time}
\vspace{-6mm}
\end{figure}





\subsection{Comparison of Trajectories}
The trajectories used in the experiment have different granularity of steps, which could impact the performance of the end model and computation time for the overall search. Therefore, we analyse the model accuracy and computation time for each step of all trajectories. As shown in Figure~\ref{fig:trajectories}, we observe that a low granularity of steps yields models with better size and WER. This could be due to smaller distortion of weights in each step, compared to weights distortion in higher granularity steps. At the same time, the overall compute time required to perform the compression process with fine grained steps does not increase significantly even though the number of steps performed is higher.
As shown in Figure~\ref{fig:trajectories_time}, this is due to the fact that even though the trajectories with bigger step sizes have fewer steps, searching for a bigger step size takes significantly longer compute time. 
Note that for a fair comparison between trajectories of different length, we keep the same overall retraining budget (i.e., 250 epochs) for all experiments, which is allocated proportionally to the granularity of steps. Therefore, the overall training time would be same for all trajectories.

\begin{figure}[t!]
    \includegraphics[width=\linewidth]{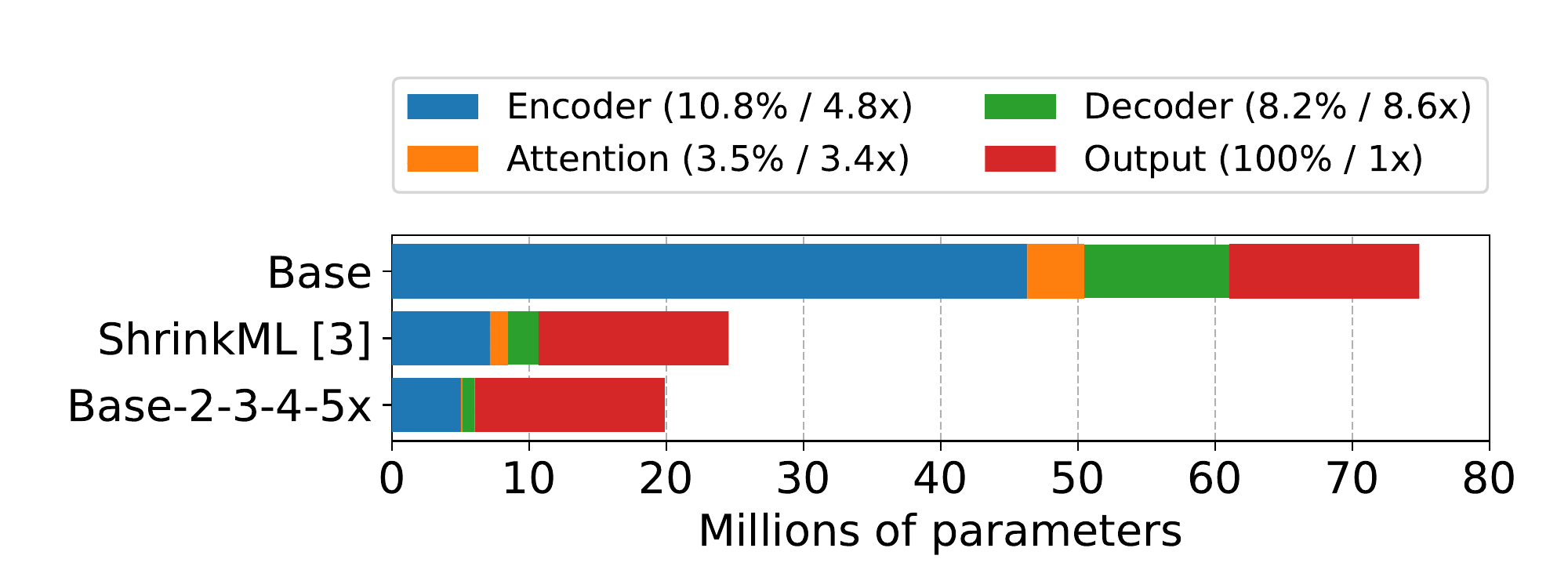}
    \vspace{-2mm}    
    \includegraphics[width=\linewidth]{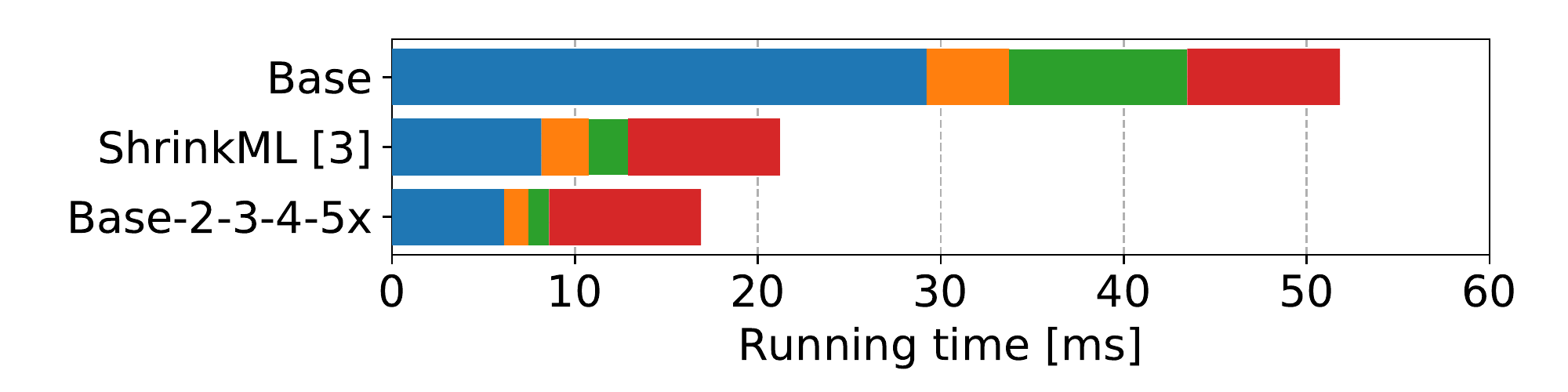}
\vspace{-4mm}
  \caption{Model compression (top) and runtime measurements on Exynos 9810 processor (bottom). Values reported in the legend are compression ratio and observed speedup of each part.}
  \label{fig:runtime}
\vspace{-4mm}
\end{figure}

\subsection{On-Device Measurements}
We compare the running-time of our 5x compressed model (with Base-2$\times$-3$\times$-4$\times$-5$\times$ trajectory) against the baseline model and 3.7x compressed model obtained by one-shot approach~\cite{shrinkml2019}. The measurements were performed on a mobile phone equipped with an Exynos 9 Octa 9810 chipset to check in practice what is the achieved speedup of our model. The models were run using a standard TFLite runtime. As shown in Fig.~\ref{fig:runtime}, improvements in running-time (bottom) remain proportional to what we would expect from theoretical estimates (top). However, it is quite visible that the speedup achieved in practice does not match exactly the theoretical one.
We attribute that to the fact that most of the layers which were compressed became very small and therefore more sensitive to any kind of overhead present in the TFLite runtime.
However, we did not study the precise reason behind this mismatch and instead leave it for future work.
\parspace

\noindent

\noindent
Overall, the running-time of the compressed model is 17ms, which indicates for the $5\times$ theoretical speedup we were able to observe only $3.0\times$ practical reduction over the baseline model. 
Compared to the current one-shot AutoML-based model~\cite{shrinkml2019}, we achieved $1.2\times$ practical reduction in running-time for the $1.35\times$ theoretical speedup.

\subsection{Why does Iterative outperform single-shot?}
To better understand differences between iterative and one-shot compression, we investigated if the one-shot compression approach fails to find the optimal ranks or the iterative compression along with retraining is a must to obtain higher compression.
To examine this, we explored two other strategies for achieving 5x speedup:
\begin{enumerate}[label=(\roman*), leftmargin=1.3em]
\vspace{-2mm}
\itemsep0em
    \item Base--5x: search for 5x speedup using one-shot AutoML approach -- i.e., analogical to~\cite{shrinkml2019};
    \item Base--Iterative Ranks: apply the best set of ranks found with our iterative approach 
    directly on the base model.
\end{enumerate}
The idea behind (ii) was to answer the question: could one-shot approach match performance of the iterative compression if it had access to a better searching method?
After compressing these models we retrained them with the same budget of 250 epochs. We used two retraining strategies: 
\begin{enumerate}[label=(\roman*), leftmargin=1.3em]
\itemsep-0.5em
    \item Compressed: in this approach we keep the model in compressed form (i.e., keeping weight matrices as $U, V$) to constrain the training to maintain the compressed size;
    \item Cyclic: in this approach we tried to mimic the training scheme from iterative compression. At every 50 epochs, we compress the model but keep the full weight matrices (i.e., as $\hat{M}$) to allow training to recover from compression. However, in the final step we train the model in the compressed form to constrain the compressed size. Moreover, after every 50 epochs the learning rate is reset to its initial value. 
\end{enumerate}

\begin{table}[t]
\centering
\begin{footnotesize}
\begin{tabular}{ l c } 
\hline
\textbf{Approach} & \textbf{WER (test-clean)}\\ 
\hline
\hline
Baseline & 8.32\\ 
\hline
Iterative Approach & 8.24 \\ 
\hline
Base--Iterative Ranks (Compressed) & 9.17 \\ 
\hline
Base--Iterative Ranks (Cyclic) & \textbf{8.19} \\ 
\hline
Base--5x (Compressed) & 10.25 \\ 
\hline
Base--5x (Cyclic) & 8.92 \\ 
\hline
\end{tabular}
\vspace{1mm}
\caption{Evaluation of different compression strategies.}
\label{table:eval}
\vspace{-6mm}
\end{footnotesize}
\end{table}

\begin{figure}[t!]
\centering
    \includegraphics[width=.8\linewidth]{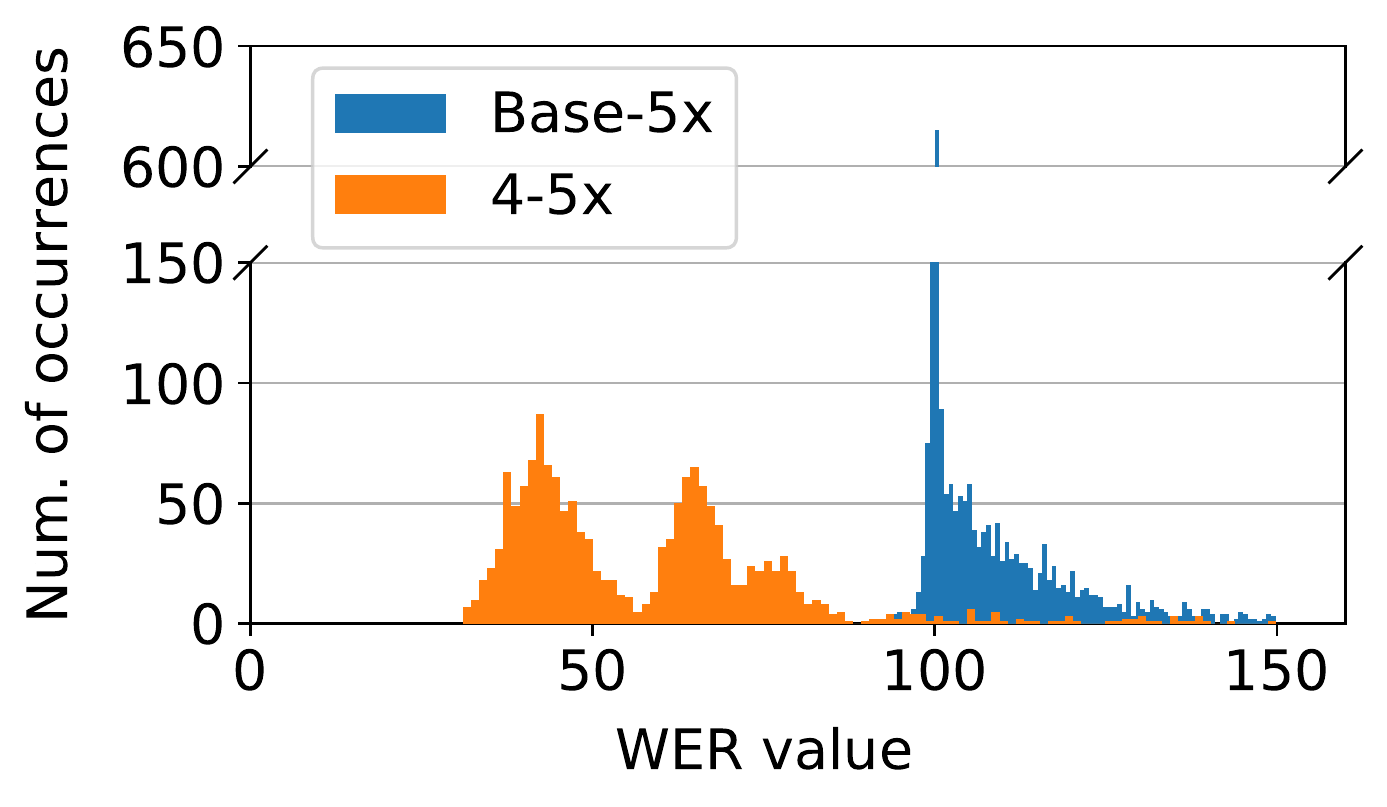}
    \vspace{-3mm}
  \caption{Iterative vs one-shot AutoML search for 5$\times$ speedup.}
  \label{fig:comparison}
\vspace{-5mm}
\end{figure}
\parspace

\noindent
As can be seen in Table~\ref{table:eval}, the results show that our iterative approach is significantly better than \textit{Base-5$\times$} approach, but when iterative ranks are applied to the baseline model (i.e., \textit{Base--Iterative Ranks} approach) we could obtain a compressed model with almost similar performance.
This indicates that there exists optimal compression schemes for large speedups that can be used to directly compress the baseline model without losing any accuracy.
%
However, as shown in Figure~\ref{fig:comparison}, the distribution of WER values for one-shot search has a strong peak, which indicates that most of the points visited by RL yield in similar rewards. This limits the algorithm to learn well as it can hardly differentiate between better and worse choices as they all become ``equally bad".

\parspace

\noindent
Furthermore, the analysis of the two \textit{Base--Iterative Ranks} strategies tells us that aggressive compression schemes, even if known to be good, can not be simply applied to the baseline model in a one-shot manner.
Instead, a different training scheme (e.g., cyclic learning rate scheduling) has to be used to fully utilize their potential, which suggests a need for more research into training techniques for small models.

%% file: conclusions.tex

In this paper we presented an AutoML-based iterative compression approach that overcomes the limitations in existing one-shot compression approaches and is found to produce highly-compressed models without any loss in WER. We applied the search on MoChA-based ASR model and demonstrated that the iterative approach could find a $5\times$ compressed model with $7\%$ relative gain in WER. 
We further showed that a baseline model can be compressed directly by using the ranks identified in the iterative search procedure. This highlights that during an one-shot search, AutoML fails to explore good ranks, since most of the compression schemes encountered by the search could not be differentiate between better and worse choices, thereby undermining the RL agent's learning ability.